\pdfoutput=1

\documentclass[11pt]{article}

\usepackage{ACL2023}

\usepackage{times}
\usepackage{latexsym}
\usepackage[T1]{fontenc}
\usepackage[utf8]{inputenc}
\usepackage{microtype}
\usepackage{inconsolata}

\usepackage{CJKutf8}

\usepackage{graphicx}
\usepackage{booktabs}
\usepackage{multirow}
\usepackage{amsmath, amsfonts, amssymb}

\newenvironment{itemize*}%
  {\begin{itemize}%
    \setlength{\itemsep}{0pt}%
    \setlength{\parskip}{0pt}}%
  {\end{itemize}}

\title{SC2: Towards Enhancing Content Preservation and Style Consistency in Long Text Style Transfer}

\author{Jie Zhao, Ziyu Guan\Thanks{~Corresponding author.}, Cai Xu, Wei Zhao \and Yue Jiang \\
        School of Computer Science and Technology, Xidian University, Xi'an, 710126, Chian\\
        \{jzhao1992@stu., zyguan@, cxu@, ywzhao@mail., 22031212489@stu.\}xidian.edu.cn}

\begin{document}
\maketitle
\begin{abstract}
Text style transfer (TST) aims to vary the style polarity of text while preserving the semantic content. Although recent advancements have demonstrated remarkable progress in short TST, it remains a relatively straightforward task with limited practical applications. The more comprehensive long TST task presents two challenges: (1) existing methods encounter difficulties in accurately evaluating content attributes in multiple words, leading to content degradation; (2) the conventional vanilla style classifier loss encounters obstacles in maintaining consistent style across multiple generated sentences.

In this paper, we propose a novel method SC2, where a multilayer Joint \textbf{S}tyle-\textbf{C}ontent Weighed (JSCW) module and a \textbf{S}tyle \textbf{C}onsistency loss are designed to address the two issues. The JSCW simultaneously assesses the amounts of style and content attributes within a token, aiming to acquire a lossless content representation and thereby enhancing content preservation. The multiple JSCW layers further progressively refine content representations. We design a style consistency loss to ensure the generated multiple sentences consistently reflect the target style polarity. Moreover, we incorporate a denoising non-autoregressive decoder to accelerate the training. We conduct plentiful experiments and the results show significant improvements of SC2 over competitive baselines. Our code: \url{https://github.com/jiezhao6/SC2}.

\end{abstract}

\section{Introduction}
\label{sec:introduction}
Text style transfer (TST) aims to generate a text exhibiting a desired style based on the source text (e.g., negative $\rightarrow$ positive), while endeavoring to faithfully preserve the semantic content.
The applications of TST cover a wide range of user-centric natural language generation tasks, such as personalized dialogue systems \cite{style_diag}, educational platforms \cite{Formality_Style}, and writing assistants \cite{styleLM}.

Given the scarcity of parallel data (i.e., text pairs conveying the same content but differing in styles) and the labor-intensive nature of annotating such pairs, existing research has predominantly focused on unsupervised TST. 
Recent contributions in this domain, including studies by \cite{word_level_relavance, reverse_attention, nast, prompt_style, style_constrained, pattern_mining}, have demonstrated significant progress.
Despite notable success, these works primarily concentrate on the transfer of a single sentence, which we call short TST. This is a relatively simple task and is difficult to apply to complex scenarios, i.e., long TST such as transferring news articles and novels. It is challenging to achieve desirable content preservation and style consistency across multiple sentences for these methods.
In a very recent study, \citet{storytrans} first concentrated on the long TST task and proposed StoryTrans, which learns content representations and fills stylistic tokens in separate stages.
While somewhat effective, challenges persist in preserving content and ensuring style consistency.

\begin{figure}[t]
	\centering
	\includegraphics[width=0.45\textwidth]{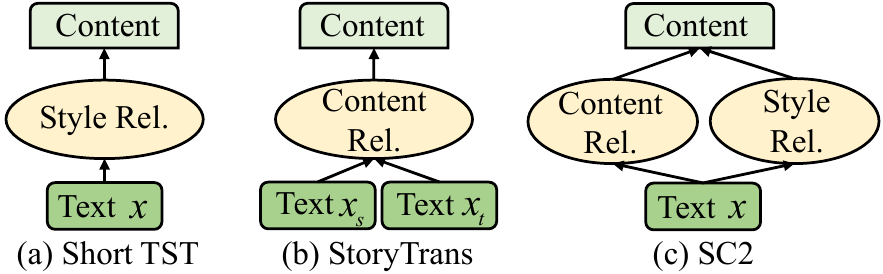}
	\caption{Comparisons of content learning between existing approaches and the proposed method: (a) evaluating the relevance between text $x$ and its style; (b) evaluating the relevance between source text $x_s$ and target text $x_t$; and (c) joint evaluating the relevance between text $x$ and its style as well as content.}
	\label{fig:content_learning}
\end{figure}

\textbf{Content Preservation}:~The critical factor for preserving content lies in accurately assessing the amount of content attribute (CA) within a token to improve the content representation. 
In traditional approaches, the content learning primarily involves explicitly \cite{frequency_mask_and_fill,nast} or implicitly \cite{word_level_relavance,reverse_attention} removing style tokens. 
In these processes, they evaluate the relevance between text and style, which solely consider the amount of style attribute (SA) within a token and neglect the CA amount (Figure~\ref{fig:content_learning}~(a)).
This results in tokens with both strong style and content attributes, such as ``euphonious''\footnote{``Euphonious'' is strongly associated with a positive sentiment style and concurrently indicates content related to music.}, potentially receiving higher SA scores, making them more drastic to be removed and consequently leading to content degradation.
On the other hand, in StoryTrans, the disentanglement of content from style is achieved by encouraging texts with distinct styles to be close together in the content space ( Firure~\ref{fig:content_learning}~(b)). However, owing to the non-parallel nature of the data, this unavoidably results in a loss of content information.

\textbf{Style Consistency}:~To control the style polarity of generated text, existing methods employ a style discriminator that operates on the entire output text. 
However, in the context of long TST, it becomes challenging for the discriminator to ensure that the style of the generated multiple sentences is consistent.
As a result, some generated sentences might exhibit strong target style polarity while others remain weak, which creates a less reader-friendly outcome. Therefore, we argue that maintaining style (polarity) consistency across multiple sentences is crucial.

To tackle the above issues, we propose a novel approach aimed at enhancing content preservation and maintaining style consistency in long TST. Our approach achieves these objectives by carefully designing a multilayer Joint \textbf{S}tyle-\textbf{C}ontent Weigher (JSCW) module and a \textbf{S}tyle \textbf{C}onsistency loss, thus we call it SC2. 
(1) The JSCW utilizes the convolution operation to measure the SA amount within the center token. 
Simultaneously, it assesses the CA amount by computing and integrating the content relevance of a token across all sentences. 
Then by normalizing these two amounts and weighting the CA amount to the tokens' representations, we obtain preliminary content representation. 
Furthermore, we employ multiple JSCW layers to progressively refine content representations. 
Finally, we fuse the target style information to content representations and feed them to the decoder to generate target text.
(2) For the other challenge, we design a contrastive learning-based style consistency loss. It brings each generated sentence closer to the previously generated sentences or target sentences in the corpus, while farther away from the sentences in the source text.

Additionally, within the unsupervised long TST setting, directly employing an autoregressive (AR) decoder substantially slows down the training since the masked self-attention technique \cite{transformer} cannot be exploited. Hence, drawing inspiration from the research on non-AR (NAR) text generation \cite{nonautoregressive,nast}, we incorporate an auxiliary denoising NAR decoder. It parallelly generates pseudo-target texts, which are then fed into the AR decoder to accelerate the training process. Our main contributions are summarized as follows:

\begin{itemize*}
    \item We propose to explicitly and simultaneously assess the SA and CA amounts within tokens to learn lossless content representations and show that this idea can significantly improve content preservation in long TST.
    \item We first propose the concept of style consistency for long TST, and design a corresponding loss to encourage the generated text to consistently maintain style polarity across multiple sentences.
    \item Extensive experiments are conducted on both Chinese and English datasets to verify the effectiveness of the proposed SC2. The results demonstrate significant improvements over competitive baselines.
\end{itemize*}

\section{Related Work}
\label{sec:related_work}

\paragraph{Text Style Transfer.}Recently, there has been considerable research attention on TST. The predominant of current efforts lies in unsupervised TST since acquiring parallel data is labor-intensive, even impossible to crowdsource for some styles \cite{survey_deep_transfer}.
One line explicitly disentangles text into content and style representations, and then incorporates the target style information into the content-based generation process. \citet{frequency_delete, frequency_mask_and_fill, reverse_attention} proposed using frequency ratio or attention scores to measure the SAs within tokens, and then entirely or partially removing the stylistic tokens. \citet{bias_mll} introduced the LIME explainer \cite{lime} to identify biased (style) tokens and then masked them for non-biased text generation. \citet{adv_disentangled} and \citet{ instance_supported_latent_space} employed the adversarial loss to learn the style-agnostic content representation. \citet{reformulating} employed a pretrained paraphrase model to strip away the style information and generated target text conditioning on the style-dependent inverse paraphrase model.
The other line uses entangled representations to perform TST. \citet{pseudo_parallel_data} and \citet{classical_and_modern_chinese} created pseudo-parallel data and then trained a \emph{Seq2Seq} model. \citet{styleLM} trained a generator for each target style. \citet{textsettr} extracted a style vector from the adjacent sentence and used it to condition the decoder.
Our work falls into the first line. Nevertheless, what distinguishes our work is the explicit and simultaneous measurement of SA and CA amounts within tokens.

The aforementioned works all handle the task of Single-Sentence (SS) transfer, conforming to an SS-input and SS-output paradigm. 
Differing from these approaches, \citet{contextual_transfer} introduced the use of context sentences as additional information to preserve topical coherence between the generated sentence and its surrounding context. Although it employs multiple sentences as input, fundamentally, this work still falls under the setting of SS transfer. 
\citet{storytrans} introduced the task of long TST and proposed a two-stage training and inference framework to learn content representations by aligning source texts and target texts. We follow this task, aiming to enhance content preservation and style consistency.

\paragraph{Non-AutoRegressive (NAR) Text Generation.}\citet{nonautoregressive} first introduced the NAR concept in machine translation which produces output tokens in parallel. Subsequent works introduce the NAR decoder into other domains, such as dialogue generation \cite{dialogue-generation}, speech synthesis \cite{text-to-speech}, and short TST \cite{nast}. 
Inspired by these works, we incorporate an auxiliary denoising NAR decoder to accelerate the training of SC2.

\section{Approach}
\label{sec:method}
\begin{figure*}[t]
    \centering
    
    \includegraphics[width=0.85\textwidth,height=0.33\textwidth]{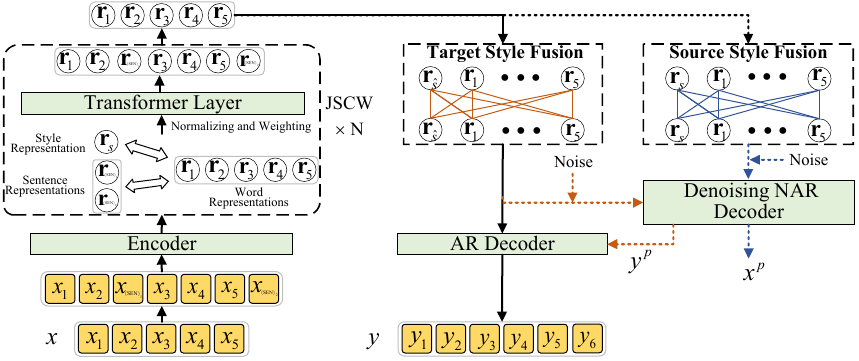}
    \caption{The framework of SC2. We show an example of transferring $x$ (long source text) to $y$ (long generated text). Dashed lines with arrows indicate that these processes occur only during the training phase.}
    \label{fig:framework}
\end{figure*}

\subsection{Problem Definition}

Let $x = (x_1, x_2, \cdots, x_{T_x})$ be a source text with multiple sentences and $s \in \mathcal{S}$ be its style label, where $\mathcal{S}$ is the style label set. 
Our goal is to develop a model $f_{\theta}$ that can generate a text $y = (y_1, y_2, \cdots, y_{T_y})$ exhibiting a desired target style $\hat{s}$ ($\hat{s} \in \mathcal{S}$ and $\hat{s} \neq s$) for a given text $x$. 
The generated text $y$ shares the same semantic content with $x$. 
Subsequently, we use bold face lower/upper case letters to denote vectors/matrices respectively.

\subsection{Model Architecture}
As shown in Figure~\ref{fig:framework}, the proposed SC2 fits within the encoder-decoder framework, and mainly consists of multilayer JSCW module, style fusion module, and denoising NAR decoder. After the encoding of original tokens, the JSCW evaluates the SA and CA amounts simultaneously and learns elementary content representations. And we propose to progressively refine content representations by multiple JSCW layers. 
Subsequently, the style fusion module combines the style information and style-agnostic content representations to provide signals of transfer direction. And the results are fed into the decoder.
Additionally, the denoising NAR decoder generates pseudo-target text to accelerate the training. 
For simplicity, our discussion in the following section focuses on the transfer direction of $s \rightarrow \hat{s}$. Nevertheless, our approach can be easily extended to handle transfers among multiple styles by incorporating multiple style embeddings.

\subsubsection{Multilayer Joint Style-Content Weigher}
\label{sec:jscw}

Assuming that the length of $x$ is $n$ and it consists of $m$ sentences, we insert a special token $\langle\texttt{SEN}\rangle$ at the end of each sentence, inspired by \cite{sentence_unshuffling, storytrans}. 
Consequently, we feed $x$ into the encoder, obtaining $n+m$ hidden states (i.e., \{$\mathbf{x}_i \in \mathbb{R}^{d}\}_{i=1}^{n+m}$). 
For convenience, we use $\mathbf{x}_{i}~(i \in [1, n])$ and $\mathbf{e}_{j}~(j \in [1, m])$ to denote the representations of $i$-th word token and $j$-th sentence token respectively. 
Inspired by the work of \cite{reverse_attention}, we assume that the representation of a word token is comprised of style information $\alpha_i \mathbf{x}_{i}$ and content information $\beta_i \mathbf{x}_{i}$, where $\alpha_i$ is the SA score (amount), $\beta_i$ is the CA score (amount), and $\alpha_i + \beta_i = 1$. The essence of the JSCW lies in discerning the values of $\alpha_i$ and $\beta_i$ simultaneously to effectively disentangle the content from style for a word token.

\paragraph{Style Weigher.}We introduce a learnable style embedding $\mathbf{s} \in \mathbb{R}^d$ to assist the computation of SA score, which will also serve as a style control signal. A straightforward idea is to directly measure the SA score based on the current word token (e.g., $\mathbf{x}_i \mathbf{W}_s \mathbf{s}$, where $\mathbf{W}_s$ is a bilinear term enabling flexible estimation). 
However, individual words sometimes do not reflect their association with a specific style. 
Therefore, we employ a convolution operation applied to a window of $2h + 1$ tokens, generating a new feature used to assess the SA score of the center token. Formally, the non-normalizing SA score of a word token is computed as follows:
\begin{equation}
\label{eq:style}
    \tilde{\beta}_i = \sigma(\mathbf{W}_{conv} \mathbf{x}_{i-h:i:i+h} + \mathbf{b}_{conv}) \mathbf{W}_s \mathbf{s},
\end{equation}
where $\sigma(x) = \text{max}(x, 0)$ is the rectifier activation function, and $\mathbf{W}_{conv}$ and $\mathbf{b}_{conv}$ are convolution matrix and bias vector respectively.

\paragraph{Content Weigher.}The content of a word token varies across diverse contexts, exemplified by the polysemy. Hence, employing a single representation akin to style embedding is inadvisable for assessing a word's CA score. 
On the other hand, in a multi-sentence scenario, a word's semantics may be interconnected with or influenced by sentences beyond its immediate sentence.
To address these challenges, we compute the content-relevance scores between a word and all sentences, subsequently attentionally aggregating them as the non-normalizing CA score. Specifically, we employ the following equation to achieve this objective:
\begin{equation}
\label{eq:content}
\begin{split}
    \mu_{i,j} &= \frac{exp(\mathbf{x}_i \mathbf{W}_c \mathbf{e}_j)} {\sum\nolimits_{j=1}^{m} exp(\mathbf{x}_i \mathbf{W}_c \mathbf{e}_j)}, \\
    \tilde{\alpha}_i &= \sum\nolimits_{j=1}^{m} \mu_{i, j} (\mathbf{x}_i \mathbf{W}_c \mathbf{e}_j),
\end{split}
\end{equation}
where $\mathbf{W}_c$ is a bilinear term.

After obtaining the individual SA score and CA score, we normalize them as: $[\alpha_i, \beta_i] = \text{softmax} ([\tilde{\alpha}_i, \tilde{\beta}_i])$.
Finally, the content representation of a word token is the weighted $\mathbf{x}_i$:
\begin{equation}
\label{eq:content_rep}
    \mathbf{x}_i^c = \beta_i \mathbf{x}_i.
\end{equation}

We define the above disentanglement process (Eqs.~(\ref{eq:style})$\sim$(\ref{eq:content_rep})) as a JSCW layer. However, the assessment of the CA score might introduce bias due to the potential inclusion of style information in the sentence representation $\mathbf{e}_j$.
To address this concern, we modify the JSCW layer by incorporating a subsequent Transformer layer \cite{transformer}.
Leveraging multi-head self-attention mechanisms, this Transformer layer utilizes elementary content representations obtained from the original JSCW layer to learn new sentence representations. 
And we employ multiple JSCW layers, which progressively diminishes the style information within sentence representations and enables a more precise assessment of CA score.

\subsubsection{Style Fusion Module}
After extracting the words' content representations, we integrate the style information into them to provide transfer signals. We concatenate the target style embedding with content representations and employ multiple Transformer layers (MTL) as the fusion network to learn style-aware representations. Formally, the process is formulated as:
\begin{equation}
    \mathbf{Z}_{\hat{s}} = \text{MTL}(Q = K = V = \tilde{\mathbf{Z}}_{\hat{s}}),
\end{equation}
where $\tilde{\mathbf{Z}}_{\hat{s}} = [\hat{\mathbf{s}}, \mathbf{x}_1^c, \mathbf{x}_1^c, \cdots, \mathbf{x}_n^c] \in \mathbb{R}^{d \times (n+1)}$, and $Q$, $K$ and $V$ respectively represent the query, key, and value within the Transformer layer. 
In summary, we denote the encoder, multiple JSCW layers, and the style fusion module as a network $F$, i.e., $\mathbf{Z}_{\hat{s}} = F(x, \hat{s})$.

\subsubsection{Decoder}
We employ an autoregressive (AR) decoder to generate target text. The decoder accesses style-aware words' representations $\mathbf{Z}_{\hat{s}}$ through the cross-attention layer. The AR decoder is defined as:
\begin{equation}
\label{eq:decoder_ar}
    P_{D_{AR}}(y \mid \mathbf{Z}_{\hat{s}}) = \prod\nolimits_{t=1}^{T_y} P_{AR}(y_t \mid y_{< t}, \mathbf{Z}_{\hat{s}})
\end{equation}

However, a practical issue arises in terms of time consumption during training. In the unsupervised setting, the decoder cannot leverage the masked self-attention technique \cite{transformer} to efficiently train the network and can only generate target tokens one by one.
For long TST, it significantly increases the training time. 
Hence, inspired by recent research on non-AR (NAR) text generation \cite{nonautoregressive, nast}, we propose a simple but effective denoising NAR decoder\footnote{The NAR decoder will be removed during inference since a simple NAR generator may not perform very well.}, which generates pseudo-target tokens in parallel, to accelerate the training. 
The generated text $y^{p}$ will be fed into the AR decoder to exploit the masked self-attention to generate target-styled text. The denoising NAR decoder is defined as:
\begin{equation}
\label{eq:decoder_nar}
\resizebox{.88\linewidth}{!}{$
    P_{D_{NAR}}(y^{p} \mid \mathbf{Z}_{\hat{s}}, \mathbf{Z}_{\hat{s}}^{n}) = \prod\nolimits_{t=1}^{T_y} P_{D_{NAR}}(y^{p}_t \mid \mathbf{Z}_{\hat{s}}, \mathbf{Z}_{\hat{s}}^{n}),
$}
\end{equation}
where $\mathbf{Z}_{\hat{s}}^{n}$ is a corrupted version of $\mathbf{Z}_{\hat{s}}$. Specifically, we randomly swap a token with one of its $2k$ neighboring tokens, where the swapping probability is $p$. We also employ a source style embedding in the fusion module and use the NAR to obtain $x^p$. During training, the NAR decoder will learn to acquire the capability of reconstructing source texts of various styles. 
It serves as providing pseudo-labels to the AR decoder, which is somewhat similar to some pseudo-parallel data construction approaches \cite{pseudo_parallel_data}. The key difference lies in the end-to-end training for both denoising NAR and AR decoders.

\subsection{Training Objectives}
\label{sec:method:training}
In this section, we elaborate on the training objectives employed in SC2.

\paragraph{Style-Oriented Objectives.} To constrain the style polarity of the generated text, we employ a document-level objective and design a sentence-level objective.
We pre-train two style classifiers $C_{docu}$ and $C_{sent}$ using standard cross-entropy loss on the entire long text and segmented sentences, respectively. The weights of these two classifiers are frozen during the training of SC2.

We first use the widely employed vanilla style transfer loss based on the document-level classifier:
\begin{equation*}
    \mathcal{L}_{sty_d} = -\mathbb{E}_{y \sim D_{AR}} [ \log P_{C_{docu}} (\hat{s} \mid y) ].
\end{equation*}

Secondly, we consider maintaining the consistency of style polarity across multiple sentences.
It is challenging for the document-level style classifier to achieve this goal.
Therefore, we derive the following contrastive style consistency loss:
\begin{equation*}
\label{eq:style_consis}
\resizebox{.98\linewidth}{!}{$
    \mathcal{L}_{sty_s} = - \frac{1}{\hat{m}} \sum\limits_{j=1}^{\hat{m}} \sum\limits_{j^{\prime}=1}^{j-1} \log (\frac{exp((\mathbf{o}_{j}^{y} \cdot \mathbf{o}_{j^{\prime}}^{y}) / \tau)} {\sum\nolimits_{j^{\prime\prime}=1}^{m} exp((\mathbf{o}_j^y \cdot \mathbf{o}_{j^{\prime\prime}}^x) / \tau)}),
$}
\end{equation*}
where $\tau$ is a temperature parameter, $\hat{m}$ ($m$) is the number of sentences in $y$ ($x$), $j$ and $j^{\prime}$ ($j^{\prime\prime}$) are the sentence indexes in $y$ ($x$), and $\mathbf{o}$ is the output of the last layer of the sentence-level classifier.
When $j=1$, the $\mathbf{o}_{j^{\prime}}^y$ is obtained by randomly sampling a target sentence from the training corpus. 
This loss makes the $j$-th generated sentence maintain stylistic similarity with sentences $1~\sim~j-1$ and target sentences in corpus, while diverging from the source sentences. This ensures both the style consistency and polarity of the generated text.

The overall loss for controlling style polarity is defined as:
\begin{equation}
\label{eq:loss_sty}
    \mathcal{L}_{sty} = \mathcal{L}_{sty_d} + \mathcal{L}_{sty_s}.
\end{equation}

\paragraph{Content-Oriented Objectives.} We employ a reconstruction loss containing self- and cycle-reconstruction, which is formulated as: 
\begin{equation}
\label{eq:loss_rec}
\begin{split}
    \mathcal{L}_{rec} = - &\mathbb{E}_{x \sim P_{x}} [ \log P_{D_{AR}} (x \mid F(x, s)) ] - \\
    &\mathbb{E}_{x \sim P_{x}} [ \log P_{D_{AR}}(x \mid F(y, s))  ].
\end{split}
\end{equation}

\paragraph{NAR Decoder-Oriented Objective.} The denoising NAR decoder is trained to reconstruct the source text, and we use the following loss:
\begin{equation}
\label{eq:loss_nar}
    \mathcal{L}_{NAR} = - \mathbb{E}_{x \sim P_x} [ \log P_{D_{NAR}} (x^p \mid \mathbf{Z}_{s}, \mathbf{Z}_{s}^{n}) ],
\end{equation}
where $\mathbf{Z}_s = F (x, s)$.

\paragraph{Disentanglement-Oriented Objective.} We empirically observe that during the early stage of training, the distribution formed by $[\alpha_i, \beta_i]$ tends to align with the Bernoulli distribution. This undermines the disentanglement learning. We design a penalizing term to mitigate this issue.
\begin{equation}
\label{eq:loss_dis}
    \mathcal{L}_{dis} = \sum\nolimits_{i=1}^{n} \max (0, \epsilon - \mathcal{H}([\alpha_i, \beta_i])),
\end{equation}
where $\mathcal{H}(\cdot)$ represents the entropy of a distribution, and $\epsilon$ is a threshold. 

In summary, the overall loss of SC2 is defined to balance the above losses:
\begin{equation}
\label{eq:loss_total}
    \mathcal{L} = \mathcal{L}_{rec} +  \lambda_{1}\mathcal{L}_{sty} +
    \lambda_{2}\mathcal{L}_{NAR} +
    \lambda_{3}\mathcal{L}_{dis}.
\end{equation}

An training issue arises with Eqs.~(\ref{eq:loss_sty})$\sim$(\ref{eq:loss_rec}), where discrete tokens in $y$ prevent the back-propagation of gradients. We address this issue by employing the soft-sampling strategy \cite{style_transformer}.

\section{Experiments}
\label{sec:exp}

\subsection{Dataset}
Following the work of \citet{storytrans}, we employ the stylized story dataset to evaluate the performance of SC2. The dataset encompasses sub-datasets in both Chinese and English. Important statistics are summarized in Table~\ref{res:dataset}.

The \textbf{Chinese} sub-dataset comprises three styles of texts: LuXun (LX)'s novels, JinYong (JY)'s novels\footnote{LuXun was a realism novelist and JinYong was known for his martial arts novels.}, and Fairy Tale (FT), totaling approximately 7.5k samples for training. The corresponding tasks involve transforming an FT into text with LX's style (FT $\rightarrow$ LX) or JY's style (FT $\rightarrow$ JY). The \textbf{English} sub-dataset contains everyday stories and fragments from Shakespeare (SP)'s plays. There are approximately 2.3k training samples. The aim is to transfer an everyday story (ER) into the style of Shakespeare's plays (ER $\rightarrow$ SP).

\subsection{Baselines}

We employ \textbf{StoryTrans} \cite{storytrans}, a two-stage framework aiming to learn content representations and filling stylistic tokens for long TST, as the strongest baseline in our experiments. Additionally, we compare SC2 with the following baselines designed for short TST: 

\noindent\textbf{Style Transformer} \cite{style_transformer} equips the power of attention mechanism within the Transformer \cite{transformer} architecture to learn entangled text representations.

\noindent\textbf{StyleLM} \cite{styleLM} fine-tunes a pre-trained language model (PLM) on a corrupted version of the style-specific corpus for each style.

\noindent\textbf{Reverse Attention} \cite{reverse_attention} employs reverse attention scores, which assess each token's contribution to the style classification, to implicitly remove the style information in tokens.

\noindent\textbf{AugZero-Shot} \cite{zero_shot} is an augmented zero-shot prompting method, which prompts the PLM with diverse sentence-rewriting examples.

\begin{table}
    \small
    \centering
    \scalebox{0.88}{
    \begin{tabular}{c|ccc|ccc}
    
    \toprule
    Dataset & \multicolumn{3}{c|}{\textbf{Chinese}} & \multicolumn{3}{c}{\textbf{English}}\\

    \midrule
    Statistics & Style & Size & Avg L & Style & Size & Avg L \\

    \midrule
    \multirow{3}{*}{Training}

    & JY & 2,964 & 344 & SP & 1,161 & 71 \\
    & LX & 3,036 & 168 & RS & 1,161 & 49 \\
    & FT & 1,456 & 175 & - & - & - \\

    \midrule
    Val. & FT & 242 & 175 & RS & 290 & 48 \\

    \midrule
    Test & FT & 729 & 176 & RS & 290 & 50 \\
    
    \bottomrule
  
    \end{tabular}}
    \caption{Statistics of the employed dataset.}
    \label{res:dataset}
\end{table}

\subsection{Evaluation Metrics}
Following previous studies of \cite{styleLM, reverse_attention, zero_shot, storytrans}, we employ both automatic and human evaluations.

\subsubsection{Automatic Evaluation}

\textbf{Style Transfer Accuracy:}~To evaluate the style polarity of the generated text, we separately pre-trained document- and sentence-level style classifiers for each language. The style classifier utilizes the same architecture as the encoder of SC2 and incorporates a mean-pooling layer and a classification layer. 
In the case of Chinese, our style classifier achieves accuracies of 100\% and 98\% for document- and sentence-level test corpus, respectively. For English, these accuracies stand at 100\% and 100\%. 
Consequently, we employ two metrics, denoted as $\text{Acc}_\text{d}$ and  $\text{Acc}_\text{s}$, to measure document- and sentence-level style transfer accuracy, respectively.

\textbf{Content Preservation:}~Firstly, we employ $\text{BLEU}_\text{1}$ and $\text{BLEU}_\text{2}$ \cite{bleu} to evaluate the content preservation through n-gram overlap. Secondly, we employ the BERT Score \cite{bert_score} metric. It leverages PLMs to assess the semantic similarity of tokens and provides a more human-like evaluation. Precision ($\text{BS}_\text{P}$), recall ($\text{BS}_\text{R}$), and F1 ($\text{BS}_\text{F1}$) scores are reported.

\textbf{Overall:}~Recognizing that content preservation and style transfer as two major challenges within the TST task, we report two metrics for the overall performance \cite{reverse_attention, storytrans} and use them as the main metrics. Specifically, we employ the geometric means of $\text{Acc}_\text{d}$ and $\text{BLEU}$ ($\text{G-BL}_\text{d}$), and $\text{Acc}_\text{d}$ and $\text{BS}_\text{F1}$ ($\text{G-BS}_\text{d}$)).

\begin{table*}[!t]
    \small
    \centering

   \scalebox{0.9}{
    \begin{tabular}{c|l|ccccccc|cc} \toprule
        Task & Methods & $\text{Acc}_{\text{d}}$ & $\text{Acc}_{\text{s}}$ & $\text{BLEU}_{\text{1}}$ & $\text{BLEU}_{\text{2}}$ & $\text{BS}_{\text{P}}$ & $\text{BS}_{\text{R}}$ & $\text{BS}_{\text{F1}}$ & $\text{G-BL}_{\text{d}}$ & $\text{G-BS}_{\text{d}}$ \\

        \midrule
        \multirow{6}{*}{\textbf{FT $\rightarrow$ LX}} 
        
        & Style Transformer & 0.1 & 1.7 & 71.7 & 64.4 & 91.7 & 92.6 & 92.1 & 3.1 & 3.6 \\
        & StyleLM   & 0.1 & 6.1 & 71.2 & 61.1 & 90.6 & 92.2 & 91.4 & 3.0 & 3.5 \\
        & Reverse Attention   & 6.4 & 28.0 & 17.3 & 4.8 & 61.4 & 58.8 & 60.0 & 8.4 & 19.7 \\
        & AugZero-Shot   & 5.9 & 38.5 & 11.7 & 4.1 & 56.5 & 59.8 & 58.0 & 6.8 & 18.5 \\
        & StoryTrans   & 39.6 & 52.0 & 25.7 & 9.6 & 63.0 & 63.9 & 63.4 & 26.4 & 50.1 \\
        \cmidrule{2-11}
        & Ours   & 41.6 & 70.5 & 27.5 & 11.9 & 61.6 & 65.5 & 63.5 & \textbf{28.6} & \textbf{51.4} \\

        \midrule
        \multirow{6}{*}{\textbf{FT $\rightarrow$ JY}} 
        
        & Style Transformer & 0.8 & 3.4 & 71.8 & 64.5 & 91.6 & 92.6 & 92.1 & 7.5 & 8.7 \\
        & StyleLM   & 1.2 & 0.9 & 70.6 & 60.3 & 90.3 & 92.1 & 91.1 & 9.0 & 10.6 \\
        & Reverse Attention   & 63.8 & 29.1 & 14.8 & 3.9 & 59.1 & 60.1 & 59.6 & 24.4 & 61.6 \\
        & AugZero-Shot   & 56.0 & 5.0 & 12.0 & 4.2 & 56.4 & 59.5 & 57.8 & 21.2 & 56.9 \\
        & StoryTrans & 70.6 & 25.1 & 22.3 & 7.9 & 61.2 & 65.5 & 63.2 & 32.7 & 66.8 \\
        \cmidrule{2-11}
        & Ours   & 74.9 & 41.5 & 26.2 & 12.5 & 64.2 & 67.7 & 65.9 & \textbf{38.1} & \textbf{70.2} \\

        \midrule
        \midrule
        \multirow{6}{*}{\textbf{ER $\rightarrow$ SP}} 
        
        & Style Transformer & 1.4 & 5.9 & 87.0 & 82.4 & 97.1 & 97.5 & 97.3 & 10.8 & 11.6 \\
        & StyleLM   & 0.3 & 2.7 & 83.6 & 80.0 & 96.0 & 97.3 & 96.7 & 5.3 & 5.8 \\
        & Reverse Attention  & 43.8 & 45.7 & 8.7 & 2.6 & 73.6 & 81.9 & 77.4 & 15.8 & 58.2 \\
        & AugZero-Shot   & 7.6 & 28.5 & 19.7 & 14.1 & 81.7 & 85.7 & 83.7 & 11.3 & 25.2 \\
        & StoryTrans & 55.9 & 47.4 & 22.0 & 7.1 & 82.1 & 83.7 & 82.9 & 28.5 & 68.0  \\
        \cmidrule{2-11}
        & Ours   & 60.3 & 54.4 & 27.3 & 10.6 & 82.7 & 84.7 & 83.6 & \textbf{33.8} & \textbf{71.0} \\

        \bottomrule
        
    \end{tabular}
    }
    \caption{Automatic evaluation results. Higher values are considered desirable for all metrics.}
    \label{exp:auto}
\end{table*}

\citet{storytrans} have noted that the perplexity based on PLMs is unreliable for fluency evaluation on this dataset. Consequently, we employ the manual evaluation to assess the fluency of the generated text.

\subsubsection{Human Evaluation}
We randomly select 100 samples from the Chinese test set and generate target texts with LX style and JY style for baselines and proposed SC2. Given the source text and the target style, three Chinese native speakers are assigned the task of scoring for style polarity (Sty), content preservation (Con), and fluency (Flu), using a scale of 1 (very bad) to 3 (very good). We report the average scores across the three annotators as our final results.

\subsection{Implementation Details}
We employ $\text{LongLM}_{\text{SMALL}}$ \cite{longlm} and $\text{T5}_{\text{SMALL}}$ \cite{t5} as backbone models for Chinese and English data respectively. For the NAR decoder, we use 6 Transformer layers. The number of JSCW layers is set to 3. For Chinese data, we set $\lambda_{1}/\lambda_{2}/\lambda_{3}$ to $0.05/1/1$, $p/k$ for the NAR decoder to $0.3/2$, the threshold $\epsilon$ to $0.15$, the batch size to 2, and the learning rate to $5 \times 10^{-5}$ (Adam optimizer). For English data, the hyper-parameters are the same except that $\lambda_{1}/p/k$ are set to $0.001/0.1/1$. The experiments are conducted on one RTX 4060Ti GPU (16G) and the training of SC2 approximately takes 7\textasciitilde9 hours.

\subsection{Results}

Table~\ref{exp:auto} shows the experimental results using automatic metrics. We can obtain the following observations: (1) the methods (StoryTrans, SC2) that are carefully designed for long TST usually perform better than traditional methods (Style Trnasformer, StyleLM, Reverse Attention, and AugZero-Shot). This demonstrates a specialized model is essential for long TST. 
(2) The Style Transformer and StyleLM exhibit significant shortcomings in controlling style polarity, while the high BLEU scores and Bert scores suggest substantial copying of the source text. This may be attributed to the difficulty of transferring entangled text representations to target tokens.
(3) Compared to StoryTrans, the proposed SC2 demonstrates significant advantages in both style polarity control and content preservation, particularly concerning the control of sentence style.
(4) The difficulties of imitating the writing styles of different authors vary. The similar BLEU metrics for SC2 suggest consistent surface-level content preservation across three sub-tasks. However, significant differences are observed in the control of document- and sentence-level style polarities for three sub-tasks.
(5) For the overall metrics, we use t-test with significance level 0.05 to test the significance of performance difference. Results show that SC2 significantly outperforms all the baselines. 

\begin{table}[!t]
    \small
    \centering

    \scalebox{0.9}{
    \begin{tabular}{c|l|ccc} \toprule
        Task & Methods & Sty & Con & Flu  \\
        
        \midrule
        \multirow{4}{*}{\textbf{FT $\rightarrow$ LX}} 
        
        & Reverse Attention   & 0.72 & 1.40 & 1.49  \\
        & AugZero-Shot   & 0.85 & 1.03 & \textbf{2.35}  \\
        & StoryTrans   & 1.87 & 1.73 & 1.50  \\
        \cmidrule{2-5}
        & Ours   & \textbf{2.29} & \textbf{1.96} & 1.85  \\

        \midrule
        \multirow{4}{*}{\textbf{FT $\rightarrow$ JY}} 
        
        & Reverse Attention   & 1.42 & 1.50 & 1.55  \\
        & AugZero-Shot   & 1.30 & 1.49 & \textbf{2.56} \\
        & StoryTrans   & 2.28 & 1.70 & 1.61  \\
        \cmidrule{2-5}
        & Ours   & \textbf{2.63} & \textbf{2.15} & 2.02  \\

        \bottomrule
        
    \end{tabular}
    }
    \caption{Human evaluation results.}
    \label{exp:human}
\end{table}

\begin{figure*}[!t]
    \centering
    \includegraphics[width=0.98\linewidth]{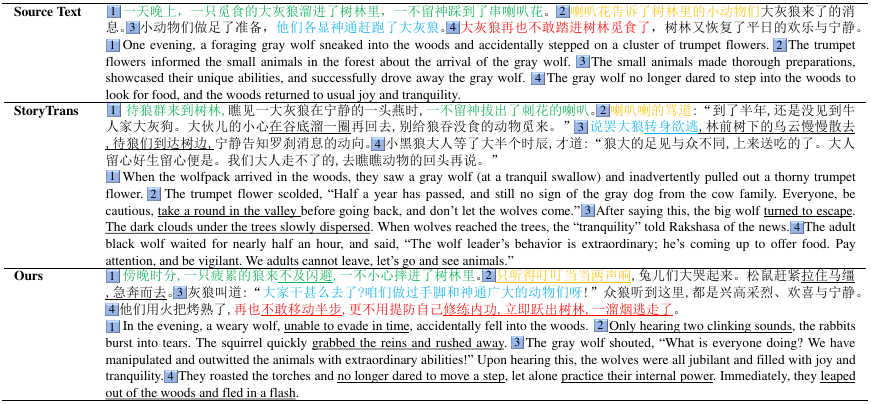}
    
    \captionsetup{type=table}
    \caption{Generated texts with the style of JY. The number preceding each sentence in the generated texts corresponds to the respective sentence in the source text in terms of semantics. Underlined sentences or phrases denote inserted contents tailored to match the target style. We use the corresponding colors of texts between the source and generated texts to emphasize the rewritten content.}
    \label{exp:case}
\end{figure*}

\begin{table*}
    \small
    \centering

    \scalebox{0.9}{
    \begin{tabular}{l|ccccccc|cc} \toprule
        Methods & $\text{Acc}_{\text{d}}$ & $\text{Acc}_{\text{s}}$ & $\text{BLEU}_{\text{1}}$ & $\text{BLEU}_{\text{2}}$ & $\text{BS}_{\text{P}}$ & $\text{BS}_{\text{R}}$ & $\text{BS}_{\text{F1}}$ &  $\text{G-BL}_{\text{d}}$ & $\text{G-BS}_{\text{d}}$ \\

        \midrule
        Ours   & 60.3 & 54.4 & 27.3 & 10.6 & 82.7 & 84.7 & 83.6 & \textbf{33.8} & \textbf{71.0} \\

        \midrule
        \multirow{3}{*}
        
        - JSCW & 57.6 & 58.1 & 20.5 & 6.0 & 81.8 & 84.1 & 82.9 & 27.6 & 69.1 \\
        -~$\mathcal{L}_{sty_s}$ & 41.7 & 46.2 & 30.2 & 12.7 & 83.6 & 85.3 & 84.4 & 29.9 & 59.4 \\

        \bottomrule
        
    \end{tabular}
    }
    \caption{Ablation study results on English dataset.}
    \label{exp:abla}
\end{table*}

As for the human evaluations, the results are shown in Table~\ref{exp:human}. Here we neglect the baselines of Style Transformer and StyleLM, which show poor performance in automatic evaluation. For content preservation and style polarity, the results from manual evaluation align with those from automatic evaluation, with our proposed SC2 achieving the highest scores. 
Additionally, annotators provided evaluations on fluency, where our proposed method significantly outperformed the StoryTrans. While the AugZero-Shot excels in fluency, its style polarity and content preservation are both too weak.
Results indicate the superiority of the proposed SC2 across various metrics.

\subsection{Analysis}

\textbf{Case Study:}~Table~\ref{exp:case} illustrates examples of generated texts with the target style of JY by the best baseline StoryTrans and SC2. In the text generated by StoryTrans, a substantial portion exhibits low relevance with the source text, indicating poor content retention. 
For style polarity, we observe that SC2 rewrites more of the source text with the target style and inserts more target-style phrases. In terms of fluency, StoryTrans performs less satisfactorily, as exemplified by the generation of peculiar phrases such as ``\begin{CJK*}{UTF8}{gkai}{在宁静的一头燕时}\end{CJK*}'' (``at a tranquil swallow''). It is evident that, in terms of content preservation, style polarity, and fluency, SC2 outperforms StoryTrans significantly.

\textbf{Ablation Study:}~We conduct ablation experiments on the English dataset to demonstrate the effectiveness of two crucial components in SC2. The results are presented in Table~\ref{exp:abla}, where ``-'' indicates that we remove the corresponding component. 
Particularly, ``-JSCW'' refers to the removal of the calculation of CA score, forming a new computational process as $\beta_{i} = \text{sigmoid}(\tilde{\beta}_{i}), \alpha_{i} = 1 - \beta_{i}$.
we can see that the performance of these variants drops apparently, which confirms the effectiveness of these proposed components in SC2.

\begin{figure}[!t]
    \centering
	\includegraphics[width=0.43\textwidth]{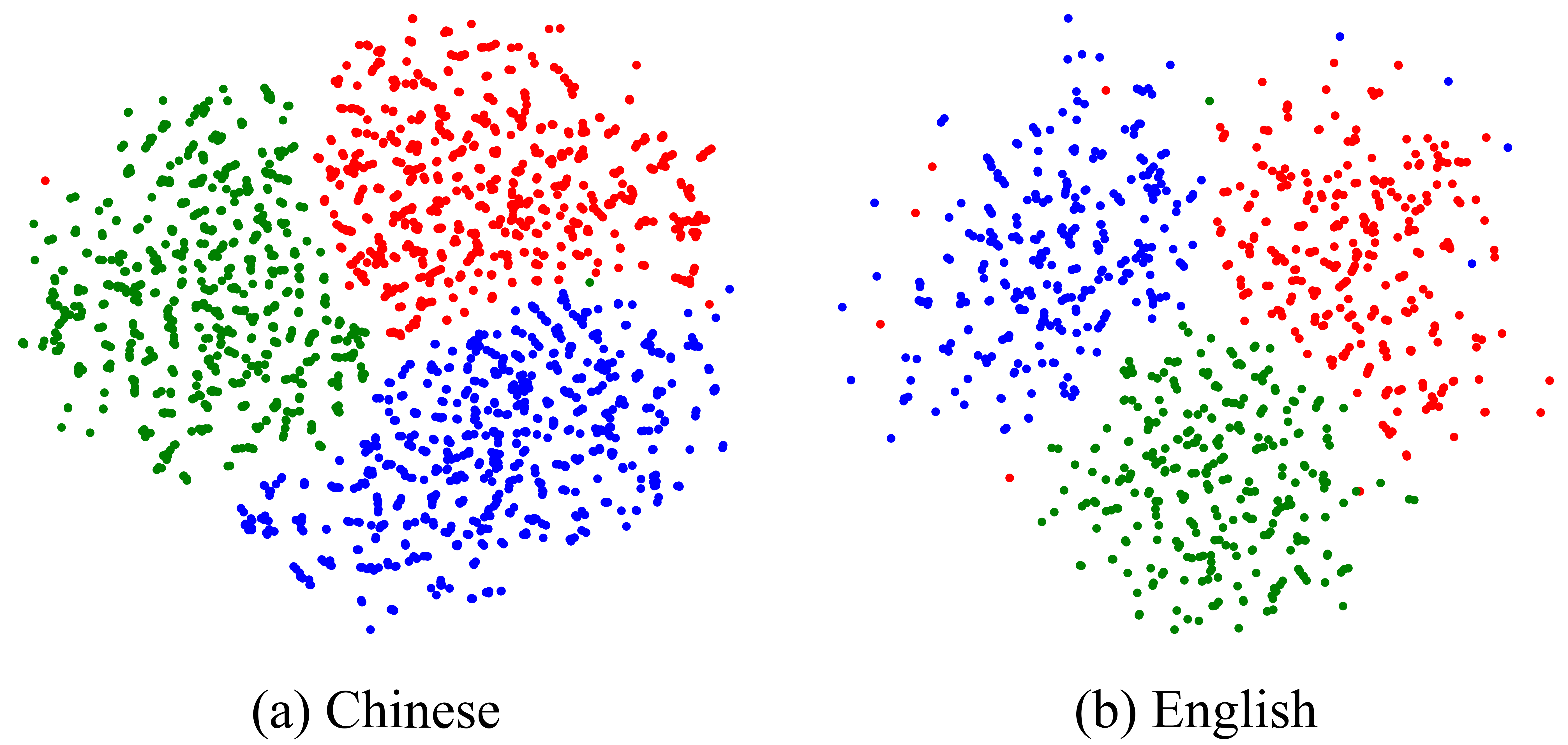}
	\caption{Visualization of Chinese and English test dataset on original space (blue dots), content space (red dots), and fused space (green dots) using t-SNE.}
	\label{exp:vis}
\end{figure}

\textbf{Visualization of the Latent Space:}~In Figure~\ref{exp:vis}, we employ the t-SNE to visualize the original, content and fused latent space of Chinese and English test datasets. 
For the Chinese dataset, we use ``JY'' as the target style. The representation of an input text $x$ is the average of all tokens. It is evident that the entangled features, disentangled content features, and fused stylistic features exhibit three distinct distributions. This demonstrates that our approach can effectively learn content information from the raw data and integrate the target style information into it.

\subsection{Extrinsic Experiment}
We further conduct an extrinsic experiment to show that the proposed SC2 can function as a data augmenter, improving the performance of data-scarce natural language processing tasks.
Specifically, we concentrate on a charge prediction task within the legal domain. Based on the factual details of legal cases, it aims at predicting the ultimate charges convicted by courts for suspects.
Traditional methods, which utilize professional legal-linguistic style (PLLS) data, typically benefit only legal experts because the domain discrepancy between PLLS and non-PLLS degrades
the models’ performance on non-PLLS texts.
However, unprofessional users also show an increasing demand on such a prediction service.
The scarcity of non-PLLS data poses significant challenges in developing effective models that benefit ordinary users \cite{dlccp}.

We address this problem by training SC2 to transfer the PLLS fact descriptions to non-PLLS texts, and then developing a charge prediction model based on the hybrid data (we mix PLLS and non-PLLS texts). The utilized datasets are CAIL \cite{cail} and NCCP \cite{dlccp} for PLLS and non-PLLS respectively.
We employ three charge prediction methods (FewShot \cite{fewshot}, CECP \cite{cecp} and DLCCP \cite{dlccp}) and three domain adaptation methods  (FADA \cite{fada}, $d$-SNE \cite{dsne} and DAGE \cite{dage}) as baselines, following the work of \cite{dlccp}. We also employ the long TST method StoryTrans \cite{storytrans} as a data augmenter.
The settings for training SC2 and StoryTrans are consistent with that of the Chinese sub-dataset in this paper. And the settings for training baselines and charge prediction model are consistent with \cite{dlccp}. We employ Accuracy (Acc.), Macro-Precision (MP), Macro-Recall (MR), and Macro-F1 (F1) as the evaluation metrics. The results are shown in Table \ref{exp:cail}. It is evident that the proposed SC2 effectively improves the performance of non-PLLS charge prediction compared to all baselines.

\begin{table}[!t]
	
	\small
	\centering

	\begin{tabular}{lcccc} \toprule

		Metrics  & Acc.  & MP    & MR    & F1 \\
		\midrule
            FewShot
            & 0.5527 & 0.5883 & 0.5677 & 0.5056 
		\\
		CECP  
		& 0.5170 & 0.5854 & 0.5461 & 0.4962 
		\\
		  FADA  
		& 0.6230 & 0.5584 & 0.6464 & 0.5482 
		\\
            $d$-SNE  
		& 0.6185 & 0.5778 & 0.6214 & 0.5528 
		\\
            DAGE  
		& 0.5959 & \underline{0.5863} & 0.6145 & 0.5473 
		\\
		DLCCP
		& \underline{0.6390} & 0.5810 & \underline{\textbf{0.6596}} & \underline{0.5619} 
		\\
        
        StoryTrans
        & 0.6094 & 0.4840 & 0.5106 & 0.4665 \\

        \midrule
        SC2
        & \textbf{0.6881} & \textbf{0.6417} & 0.6259 & \textbf{0.5860} \\

	\bottomrule
	\end{tabular}
	\caption{Extrinsic experiment results. Underlined values denote the optimal results of baselines.}
 \label{exp:cail}
\end{table}

\section{Conclusion}
\label{sec:conclusion}
In this paper, we propose a novel method SC2 for long TST. We design multiple JSCW layers to progressively refine content representations of tokens, thus enhancing content preservation. And we design a style consistency loss to ensure the generated sentences share similarities in style polarity. Experimental results with automatic and human evaluations confirm the effectiveness of SC2 compared to competitive baselines.

\section*{Limitations}
We have not yet evaluated the performance of SC2 with larger pre-trained language models since they are not mainstream methods for TST and they exceed the capacity of our available GPU resources. 
Another limitation is that, for both Chinese and English, the data used in our experiments only involves a single type of long TST, specifically imitating the writing style of a particular author. We also find the styles to be quite distinct because of the high accuracy scores obtained by style classifiers.
This limitation arises from the fact that there is only one available dataset for the long TST in the existing research. We leave the construction of diverse data (e.g., formality or humor transfer, and texts with closer styles) and the comprehensive validation of SC2's generalizability and adaptability to future work.

\section*{Ethics Statement}
Long TST involves generating target text based on both the original text and target style signals. The nature of text generation and the diversified stylistic corpus raise potential ethical concerns, including the risk of generating text that may be inappropriate, offensive, or biased. 
On the other hand, the model carries the risk of being maliciously exploited, such as generating fake news and fabricating political statements. Future research on how to mitigate these issues is in crucial need.

We hired three Chinese native speakers as annotators to manually evaluate the performance of the proposed method and baselines. Considering the wage standards of China, annotators will get 2.0 yuan (RMB) for each sample.

\section*{Acknowledgements}
This research was supported by the National Natural Science Foundation of China (Grant Nos. 62133012, 61936006, 62103314, 62073255, 62303366), the Key Research and Development Program of Shaanxi (Program No. 2020ZDLGY04-07), Innovation Capability Support Program of Shaanxi (Program No. 2021TD-05) and Natural Science Basic Research Program of Shaanxi under Grant No.2023-JC-QN-0648.

\bibliographystyle{acl_natbib}
\bibliography{custom}

\end{document}